\documentclass[letterpaper, 10 pt, conference]{ieeeconf}  % Comment this line out if you need a4paper

\IEEEoverridecommandlockouts                              % This command is only needed if 
\usepackage{graphics} % for pdf, bitmapped graphics files
\usepackage{epsfig} % for postscript graphics files
\usepackage{times} % assumes new font selection scheme installed
\usepackage{amsmath} % assumes amsmath package installed
\usepackage{amssymb}  % assumes amsmath package installed
\usepackage{amsfonts}
\usepackage{dsfont}
\usepackage{booktabs}
\usepackage{multirow}
\usepackage{algorithm}
\usepackage{algorithmic}
\usepackage{fancyhdr}

\title{\LARGE \bf
Enhanced Prototypical Learning for Unsupervised Domain Adaptation\\in LiDAR Semantic Segmentation
}

\author{Eojindl Yi$^{1}$, JuYoung Yang$^{1}$ and Junmo Kim$^{1}$% <-this % stops a space
\thanks{$^{1}$Korea Advanced Institute of Science and Technology (KAIST), Korea
{\tt\small \{djwld93, yjy6711, junmo.kim\}@kaist.ac.kr}}
}

\fancypagestyle{FirstPage}{
\lfoot{\fontsize{7pt}{7pt}\selectfont© 2022 IEEE.  Personal use of this material is permitted.  Permission from IEEE must be obtained for all other uses, in any current or future media, including reprinting/republishing this material for advertising or promotional purposes, creating new collective works, for resale or redistribution to servers or lists, or reuse of any copyrighted component of this work in other works.} 
}

\begin{document}

\maketitle
\thispagestyle{empty}
\pagestyle{empty}

%%%%%%%%%%%%%%%%%%%%%%%%%%%%%%%%%%%%%%%%%%%%%%%%%%%%%%%%%%%%%%%%%%%%%%%%%%%%%%%%
\begin{abstract}
Despite its importance, unsupervised domain adaptation (UDA) on LiDAR semantic segmentation is a task that has not received much attention from the research community. Only recently, a completion-based 3D method has been proposed to tackle the problem and formally set up the adaptive scenarios. However, the proposed pipeline is complex, voxel-based and requires multi-stage inference, which inhibits it for real-time inference. We propose a range image-based, effective and efficient method for solving UDA on LiDAR segmentation. The method exploits class prototypes from the source domain to pseudo label target domain pixels, which is a research direction showing good performance in UDA for natural image semantic segmentation. Applying such approaches to LiDAR scans has not been considered because of the severe domain shift and lack of pre-trained feature extractor that is unavailable in the LiDAR segmentation setup. However, we show that proper strategies, including reconstruction-based pre-training, enhanced prototypes, and selective pseudo labeling based on distance to prototypes, is sufficient enough to enable the use of prototypical approaches. We evaluate the performance of our method on the recently proposed LiDAR segmentation UDA scenarios. Our method achieves remarkable performance among contemporary methods.
\end{abstract}
%%%%%%%%%%%%%%%%%%%%%%%%%%%%%%%%%%%%%%%%%%%%%%%%%%%%%%%%%%%%%%%%%%%%%%%%%%%%%%%%
\section{Introduction}

\thispagestyle{FirstPage}

Recent advances in data processing and sensor development are accelerating the advent of autonomous vehicles. Perceiving the world is an important issue for an autonomous vehicle, and using LiDAR sensors is an appealing choice, because of the accurate depth information it provides. Research is now moving on towards directly extracting semantic information from the raw LiDAR data. Because deep learning and neural networks have shown high potential in learning to extract meaningful information from data across different modalities, processing LiDAR data with neural networks has also become an active research area. Advances in sensor technology and the release of large-scale LiDAR datasets are nourishing the research area even further. 

However, the problem of domain shift in LiDAR data is a significant yet prevalent problem that needs attention from the research community. Domain shift is a problem that the trained model fails to generalize because of the change in the data distribution. For natural images, one common situation is that a model trained to segment sunny weather images only may fail to generalize on cloudy weather images. Regarding LiDAR data, the difference between LiDAR sensor specs and even subtleties such as sensor displacements can result in data distribution differences. The problem needs to be addressed because, on the one hand, every time a new data sample is obtained, tremendous labeling cost and effort is required. Each LiDAR scan contains tens of thousands of points that need to be labeled. Although sensor fusion with other modalities such as camera or radar can mitigate the difficulty of the labelling process, LiDAR labeling requires supervision from skilled human resources.

On the other hand, the problem needs attention because due to the severeness of domain shift across datasets, the general benefits of the data-driven nature of deep learning do not hold. For example, in tasks that process 2D natural images, it is common to use models that are pre-trained using ImageNet~\cite{ILSVRC15} datasets. This is because, training models with more and diverse images as possible will contribute to the increased generalization capability and task performance of the model. However, due to the severeness of domain shift, such desiderata cannot be expected in LiDAR data. Because of these reasons, addressing LiDAR domain shift, and furthermore generalization is a research direction of significant importance. Among such research attempts, unsupervised domain adaptation (UDA) is a research area that, given a labeled source dataset and unlabeled target dataset from different domains, tries to correctly guess the target labels with access only to the source labels only.

Recently, \cite{yi2021complete} came up with a multi-staged method called Complete \& Label that directly tackles the UDA between LiDAR datasets. They interpreted that the domain shift results from different sampling patterns of the 3D world, and first trained a voxel-based completion network to reconstruct 3D surfaces from an input LiDAR scan. The reconstructed representation is called canonical domain, implying a dense representation regardless of the sampling pattern. Next, an additional network was trained to segment the scan from the canonical domain.

Although, Complete \& Label is a promising baseline in terms of performance, the expandability of the method is limited. Voxel-based methods generally suffer from large memory consumption and long inference time, meaning that they will be difficult to use in real-time applications requiring inference time of less than 100 ms per scan, or in conjunction with other downstream tasks such as Semantic SLAM~\cite{chen2019suma++}. Range image-based methods, which project the LiDAR scans on 2D images, satisfy such conditions but suffer from a low upper bound of performance. This necessitates the development for a method that is comparable in performance but light and fast enough to be used in realistic applications.

In this paper, we propose an effective and efficient method that solves the UDA problem in the LiDAR semantic segmentation task. Because our method is based on using 2D range images, it is free from the said complexity issues. Plus, its performance bound improves largely upon contemporary 2D methods, with the help of our method design. Our method utilizes source data prototypes to pseudo label target pixels, and reduces the domain gap by reducing the difference between source prototypes and pseudo labeled target features. The source prototypes are enhanced with the additional use of encoder features and moving averages, which are design choices tailored for the nature of LiDAR segmentation and UDA. Target pseudo labeling during training is done selectively, by only trusting a small portion of target pixels based on their distance to prototypes, whose portion increases during training. Unlike natural image segmentation, we do not have access to a pre-trained feature extraction model, which decreases the feasibility of prototypical approaches. To overcome this problem, our overall framework opts a two-staged training strategy that starts with a pre-training step that trains the model with a label-agnostic reconstruction objective, and a joint training step utilizing source labels. Benchmark performance and ablation studies support the validity of our model design. Our method exhibits remarkable performance among contemporary 2D and 3D methods.

\section{Related Work}

\subsection{Unsupervised domain adaptation methods}
In the computer vision community, many UDA methods have been developed especially on the image classification task. \cite{ganin2015unsupervised} trained a network that forgets to correctly classify the different domains. \cite{long2015learning, kang2019contrastive, zhang2019bridging} defined a measure that quantifies the divergence between the domains, and trained their networks to reduce the measure. \cite{sun2016deep, morerio2018minimalentropy} tried to reduce the difference of second order statistics between domains.

In recent years, prototypical methods~\cite{pan2019transferrable, zhang2021prototypical} have emerged, showing impressive performance. The known source labels are used to obtain class prototypes, which are in turn used for finding reliable target pseudo labels. Compared to former prototypical UDA methods, our method tries to tackle an extreme problem in which (i) due to the domain shift and the lack of pre-trained feature extractor, prototypical pseudo labels are less accurate (ii) the dataset size limits the use of calculating the source prototypes all over the dataset (iii) the representation size does not allow the use of large dimensional prototypes. In the forthcoming sections we explain and validate how each of the problems are solved.

\subsection{Unsupervised domain adaptation methods for LiDAR semantic segmentation}
In this section, we introduce UDA methods for LiDAR semantic segmentation. \cite{Wu2019SqueezeSegV2IM} addressed the UDA from simulation to real datasets. They proposed a new architecture and an extra network that learns to render the LiDAR intensity values for the simulated dataset. They also used geodesic correlation alignment~\cite{morerio2018minimalentropy} to reduce domain gap. \cite{jaritz2020xmuda} used both natural images and voxelized point clouds as input, and performed cross-modal training. They experimented on diverse realistic scenarios including day-to-night, cross-country and cross-dataset scenarios. \cite{langer2020transfer} proposed a method that generates a semi-synthetic dataset using the method from \cite{morerio2018minimalentropy} mentioned above. \cite{jiang2020lidarnet} used a multi-branch network to train the network by discerning domain-wise private and shared features. Recently, Complete \& Label~\cite{yi2021complete} proposed a new benchmark on LiDAR UDA and a completion based method to deal with the different sampling patterns of different sensors. \cite{rochan2021unsupervised} came up with a range image-based method that tackles part of the scenario proposed by \cite{yi2021complete} with the help of self-supervised auxiliary tasks. Our method outperforms \cite{yi2021complete, rochan2021unsupervised} in performance, while using a simple method and operating at a runtime that is suited for real-time applications.

\section{Method}
We propose an effective and efficient 2D projection-based method for solving the UDA problem in LiDAR semantic segmentation. Our method can be summarized as reducing the domain gap by reducing the difference between known source prototypes and pseudo labeled target features. First, the preliminaries on the data preparation, our principle for data processing and pre-training step are introduced. Next, we explain how to obtain and update the enhanced version of our prototypes. Lastly, we explain the pixel-wise pseudo labeling process on the target samples,  filtering of pseudo labels, and confidence-based weighting criterion.

\subsection{Range Data and the Source First Principle}

We first provide preliminary explanations for converting a 3D LiDAR scan to a 2D range image. Given an input LiDAR scan containing points $p = (x, y, z)$, the points are converted to a 2D range image using the following projection formula:
\begin{align}
\left(\begin{array}{c}
u \\
v
\end{array}\right)=\left(\begin{array}{c}
\frac{1}{2}\left[1-\arctan (y, x) \pi^{-1}\right] w \\
{\left[1-\left(\arcsin \left(z r^{-1}\right)+f_{\mathrm{up}}\right) \mathrm{f}^{-1}\right] h}
\end{array}\right)
\end{align}

\noindent where $u$, $v$ are the image 2D coordinates and $w$, $h$ correspond to the desired width and height resolution of the range image. $h$ is usually selected as the number of LiDAR beams, and $w$ is usually selected in relation to the angular resolution of the sensor specs. Commonly used $(h, w)$ values are $(32, 1024)$ and $(64, 2048)$, and larger image resolutions are usually preferred because of better performance. Note that we only used the point coordinates, and did not use any additional features such as intensity or RGB values.

\begin{figure*}[!t]
\centering
\includegraphics[width=1\textwidth]{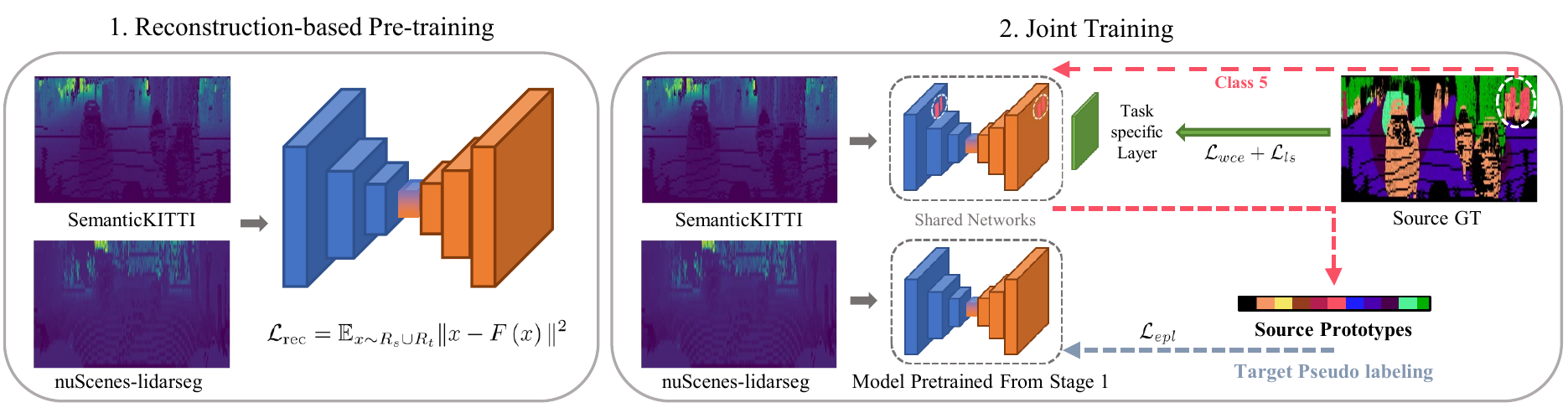}
\caption{Overall outline of our method. First, we start with a task-agnostic reconstruction pre-training using the range images from both domains. Next, we use the pre-trained model and source labels to extract source prototypes. Then we utilize the prototypes to pseudo label target pixels, and selectively train a small portion of the target pixel features to become similar to the prototypes corresponding to their pseudo labels. Meanwhile, a task specific layer is directly trained using the source labels. Because the target features are aligned with source prototypes, the difference between domains is reduced. After training has finished, the target data can be segmented using the same task specific layer.}
\label{fig:method}
\vspace{-2.5mm}
\end{figure*}

However, one drawback of 2D range projection is that, if $h$ is larger than the number of input LiDAR beams, empty pixels forming a striped pattern appear at the converted range image. This can be problematic in the LiDAR UDA setup because different LiDAR datasets are taken from different sensors, and if they are unified with a single, incompatible $(h, w)$, unwanted byproducts are guaranteed to appear. Previous work has attempted to circumvent such unwanted byproducts via upsampling or hole filling algorithms, but such rule-based strategies are at the risk of contaminating the data.

\begin{table}[t]
\centering
\renewcommand{\arraystretch}{0.95}
\caption{Details on the Source First Principle. K stands for SemanticKITTI \cite{behley2019semantickitti} and N stands for nuScenes-lidarseg \cite{caesar2020nuscenes}}
\resizebox{\columnwidth}{!}{
\begin{tabular}{c|c|c|c}
\toprule
\multirow{2}{*}{Scenario}           & \multirow{2}{*}{\shortstack{Best Input Setup\\ for Source Domain}}  & \multirow{2}{*}{Target Processing}            & \multirow{2}{*}{Network Channel} \\
                                   &                                   &                                               &                                  \\
\midrule
\midrule
\multirow{2}{*}{K2N}     & \multirow{2}{*}{(64, 2048)}       & \multirow{2}{*}{\shortstack{Upsampling\\ \& Pooling}} & \multirow{2}{*}{32}              \\
                                   &                                    &                                               &                                  \\
\cmidrule{1-4}
\multirow{2}{*}{N2K} & \multirow{2}{*}{(32, 1024)}       & \multirow{2}{*}{\shortstack{Project with \\ $(32, 1024)$}}   & \multirow{2}{*}{128}             \\
                                   &                                    &                                               &                                 \\
\bottomrule
\end{tabular}
}

\label{tab:SFprinciple}
\vspace{-2.5mm}
\end{table}

To minimize such risks as much as possible, we employ a simple strategy named \textbf{the Source First Principle}. As the name implies, we prioritize the source data since it is the only representation in our setup that can be fully trusted. Therefore, we choose the best scenario of $w$ and $h$ that best suits the source domain sensor specs, and upsample or re-project the target data to match that scenario. Details can be found on Table 1, in which we show the applied principle for two major LiDAR datasets, SemanticKITTI \cite{behley2019semantickitti} and nuScenes-lidarseg \cite{caesar2020nuscenes}. In the K2N scenario, we perform a pre-processing step based on nearest neighbor upsampling to upsample the target data representation from $(32, 1024)$ to $(64, 2048)$. During inference of the target data, we perform average pooling and reduce the resolution back to $(32, 1024)$. On the N2K scenario, the procedure is less complicated as projecting the SemanticKITTI data with (1) at a resolution of $(32, 1024)$ is sufficient enough. Because the volume of data being processed is different for each scenario, we correspondingly increase the model channel dimension to match the amount of information that is processed in every scenario.

Based on the obtained data representation, we explain the model architecture and pre-training process that is crucial for prototypical learning. Prototypical self-training in natural image segmentation is a blessing from ImageNet pre-training. Because pre-training equips the model with basic feature extraction capabilities, prototypes conditioned on source labels can act as anchors to discern reliable pseudo labels from the target domain. However, due to the domain shift in LiDAR data, pre-training and prototypical methods altogether have not been attempted in the LiDAR setup. Surprisingly, we found out that a simple task-agnostic pre-training step empowers the model enough such that cross-domain prototypes can be aligned. For a given number of epochs, we train a model with the following auto-encoder objective:
\begin{align}
\mathcal{L_{\mathrm{rec}}}=\mathbb{E}_{x\sim{R_s}\cup{R_t}}{\|{x} - {F}\left(x\right)\|^2}
\end{align}
where $x$ denotes the sampled range image from the source and target domain range image set, $R_s$ and $R_t$. $F$ denotes a task-agnostic neural autoencoder. Throughout our method, we use the SalsaNext \cite{cortinhal2021salsanext} architecture, which has shown good performance on range image-based LiDAR segmentation. $F$ is obtained by removing the task-specific fully-connected layers from SalsaNext, and the default channel of the architecture is set according to Table 1.

\subsection{Enhanced and Averaged Prototypes}
We aggregate the source data labels and features to obtain source prototypes, which can be used as anchors for target feature alignment. Similar to concurrent work~\cite{pan2019transferrable, zhang2021prototypical}, the source prototypes $\mu_{c}^{s}$ are calculated by the following equation:
\begin{align}
\mu_{c}^{s} = \sum_{n} \sum_{h} \sum_{w} \mathds{1}_{\mathbf{y}_{n,h,w}^{s} = c}
 \frac{F(\mathbf{x}_{n,h,w}^{s} )}
{\| F(\mathbf{x}_{n,h,w}^{s} ) \|}
\end{align}
where $\mathbf{x}$ is a range image, the superscript denotes the domain it is sampled from, and the three subscripts are the sample-wise index over the dataset, and the row and column indices, respectively. $c \in \{0, \ldots, C\}$ denotes the class index and $\mathds{1}$ is an indicator variable denoting $1$ if the subscript holds and $0$ otherwise. This allows us to obtain target domain prototypical pseudo labels $\hat{\mathbf{y}}_{n,h,w}^{t, pr}$ using:
\begin{align}
& sim(F(\mathbf{x}_{n,h,w}^{t} ), \mu_{c}^{s}) = \frac{\langle F(\mathbf{x}_{n,h,w}^{t} ), \mu_{c}^{s} \rangle}
{\| F(\mathbf{x}_{n,h,w}^{t} ) \|\| \mu_{c}^{s} \|}\\
& \hat{\mathbf{y}}_{n,h,w}^{t, pr} = 
\arg\max _{c} sim(F(\mathbf{x}_{n,h,w}^{t} ), \mu_{c}^{s})
\end{align}
where $sim$ is the abbreviation for similarity, $pr$ is the abbreviation of prototype and $n$ denotes sample-wise index over the dataset. $0$ corresponds to the ignored class, usually the background and the pixels with no points projected. Note that we do not ignore class $0$ in both target data and source data, which is different from the former work. This is because, if the class $0$ representations and prototypes are not taken into account, the target data class $0$ representations all collapse to the majority class. In the K2N scenario, this would mean that the background and noise pixels of nuScenes, even the easiest ones, would all be classified as the source data majority class, in this case, vegetation. This is a catastrophic situation that we naturally want to avoid.

Directly adapting the above formulation of prototypes for the segmentation task entails difficulties in the LiDAR segmentation setup. Due to the large size of input, our model capacity is limited in size, thus having a low number of channel dimension. Since our prototypes are obtained by averaging over the row and column indices $h$ and $w$, its size equals the number of channel dimensions, which means that our prototypes are generally small in size and the representative power is generally weak.

To compensate this weakness, we pay attention to the general architectural structures of segmentation networks. Usually it downsamples and upsamples a given input, meaning that a corresponding feature map $F'(\mathbf{x}_{n,h,w}^{s})$ whose resolution equals that of the final feature map $F(\mathbf{x}_{n,h,w}^{s})$ from which we average and obtain the prototypes, can be extracted from the encoder part of the network. Plugging that feature map into (3) we obtain enhanced prototypes $\widehat{\mu}_{c}^{s}$:
\begin{align}
& \mu_{c}^{'s} = \sum_{n} \sum_{h} \sum_{w} \mathds{1}_{\mathbf{y}_{n,h,w}^{s} = c}
 \frac{F'(\mathbf{x}_{n,h,w}^{s} )}
{\| F'(\mathbf{x}_{n,h,w}^{s} ) \|}\\
& \widehat{\mu}_{c}^{s} = [\mu_{c}^{'s};\mu_{c}^{s}]
\end{align}
whose representative power is increased with the help of encoder features. By $[;]$ we denote that we have performed concatenation on the channel dimension.

An additional difficulty is that the LiDAR datasets are usually large, and it is expensive to iterate over the whole dataset to obtain class prototypes. An alternative approach would be to sample source and target data at every iteration, obtain the source prototypes on the fly and obtain the target pseudo labels from those prototypes. However, this approach still fails as there is no guarantee that the source classes will appear on the target scene, and vice-versa. Therefore, we propose to keep exponential moving averages of prototypes from every class using: 
\begin{align}
\begin{split}
\widehat{\mu}_{c}^{current} = \sum_{n_{B}} &\sum_{h} \sum_{w} \mathds{1}_{\mathbf{y}_{n_{B},h,w}^{s}=c}\\ &\left[\frac{F'(\mathbf{x}_{n_{B},h,w}^{s} )}
{\| F'(\mathbf{x}_{n_{B},h,w}^{s} ) \|} ; \frac{F(\mathbf{x}_{n_{B},h,w}^{s} )}
{\| F(\mathbf{x}_{n_{B},h,w}^{s} ) \|}\right]
\end{split}\\
\widehat{\mu}_{c}^{s,i} \leftarrow \alpha\widehat{\mu}_{c}^{s,i-1}& + \left(1-\alpha\right)\widehat{\mu}_{c}^{s,current} 
\end{align}
where $i$ denotes the index for the training iteration, and $\widehat{\mu}_{c}^{s,current}$ is calculated by (6) over the sampled source batch dataset, indexed by $n_{B}$. Note that we keep moving averages of the enhanced prototypes. During training, we plug $\widehat{\mu}_{c}^{s,i}$ into (4) and (5) to obtain the target pseudo labels.

\subsection{Pseudo Label Filtering and Confidence Weighting}
The lack of pre-trained feature extractor, and the severeness of domain shift are factors that degrade the accuracy of prototypical pseudo labeling. Although we have included a pre-training step to mitigate such shortcomings, the quality of features it provides will be still inadequate, compared to the ImageNet pre-trained features used in natural images. This necessitates a more cautious strategy in deploying the pseudo labels for network training, and we propose to filter pseudo labels based on their similarity to prototypes.

For every pixel in the sampled target batch data, we calculate the maximum class similarity using (4). Our network is only trained with the filtered pixels with the top $p_{pc}$ percentile of the class-wise maximum similarities, where $p_{pc}$ is a hyperparameter obtained by multiplying the epoch with a pre-defined per epoch portion increment $p_{inc}$. Setting $\tau_{c}$ as the top $p_{pc}$ percentile value of similarities for class $c$ we define a filtering indicator as:
\begin{align}
    M_{n,h,w}^{t} =
  \begin{cases}
    1 & \text{if } \max _{c} sim(F(\mathbf{x}_{n,h,w}^{t} ), \mu_{c}^{s}) > \tau_{c}\\
    0 & \text{otherwise}
  \end{cases}
\end{align}
Since $\tau_{c}$ is a value ranging from $0$ to $1$, it can be also thought of as a class-wise confidence value for the current predictions. As we do not have accurate information on the target domain, it is best to first focus more on the accurate classes, and then progressively train on the harder classes. Therefore, we incorporate the $\tau_{c}$ values into our final loss function as a weighting term.

\subsection{Mask Deactivation and Background Down-weighting}

The datasets used in our adaptation scenarios do not contain the same set of classes, and therefore we have to ignore the non-overlapping classes. We map such classes to class $0$, but then, class $0$ becomes a mixture of the classes to ignore, background, noise, and even empty pixels on which no points are projected. Under such circumstances, ignoring class $0$ during training is an unnatural design choice, which we have pointed out in Section B. However, because class $0$ is the majority class, the training dynamics is expected to deteriorate as it becomes difficult to extract semantic representations from the majority class. To mitigate such issues we propose two solutions, mask deactivation and background down-weighting.

Originally, \cite{milioto2019rangenet} generated range images using (1) and applied normalization step based on mean subtraction and standard deviation division. Thereafter, the images were multiplied by a binary mask, which contains $1$ if any point is projected to the corresponding pixel and $0$ if no point is projected. If the original range image before normalization contained pixels with no projection their values would have become negative due to mean subtraction, and thus would have stood out compared to the surrounding range values. However, the process of multiplying the masks can be thought as dampening the apparent, easy-to-recognize regions of class $0$. To let the model recognize apparent cases of class $0$ with ease, we opt to deactivate the multiplication of binary masks. 

In general, class $0$ pixels are easy to learn, which is further eased due to the above mask deactivation process. If we obtain class prototypes and target pseudo labels, class $0$ is the most accurate and dominant one, which makes the model excessively focus on class $0$ during training. To counteract such situations, we enforce a stricter condition on class $0$ by decreasing the filtering percentile value $p_{pc}$ of class $0$. Therefore, we propose to use a $p_{0}$ percentile value for class $0$ which is obtained by multiplying $p_{pc}$ and a down-weighting term $p_{dw}$.
\begin{table}[t!]
\vspace{-4mm}
\begin{algorithm}[H]
\small
\caption{Enhanced Prototypical Learning}
\label{alg:algorithm}
\textbf{Input}: Dataset $\left\{\left(\mathbf{x}_{i,:,:}^{s},  \mathbf{y}_{i,:,:}^{s}\right)\right\}_{i=1}^{n_{s}}$, and  $\left\{\left(\mathbf{x}_{i,:,:}^{t}\right)\right\}_{i=1}^{n_{t}}$

\textbf{Parameter}: $max\_epoch, max\_iter, \alpha, \lambda$

\begin{algorithmic}[]
\STATE Let $epoch=1$
\WHILE{$ epoch \leq max\_epoch $}
\STATE Update $p_{pc}$ used in calculating $\tau_{c}$ in (10)
\STATE Let $iter=0$
\WHILE{$ iter < max\_iter $}
\STATE Forward $\mathbf{x}_{i,:,:}^{s}$ and obtain $\widehat{\mu}_{c}^{current}$ using $\mathbf{y}_{i,:,:}^{s}$ and (7)
\STATE Use $\widehat{\mu}_{c}^{current}$ to obtain $\widehat{\mu}_{c}^{s,iter}$ by updating (9)
\STATE Forward $\mathbf{x}_{i,:,:}^{t}$ and obtain $M_{i,:,:}^{t}$ using $\widehat{\mu}_{c}^{s,iter}$ and (10)
\STATE Train the network with (11), (12)
\STATE $ iter \gets iter + 1$
\ENDWHILE
\STATE $ epoch \gets epoch + 1$
\ENDWHILE
\end{algorithmic}
\textbf{Output}: Domain adapted $F$ and $G$
\end{algorithm}
\vspace{-6mm}
\end{table}

\subsection{Final loss function}
Integrating all the components of our method, the final loss will be as follows:
\begin{align}
&\mathcal{L} = \mathcal{L}_{wce} + \mathcal{L}_{ls} + \lambda \mathcal{L}_{epl}\\
\begin{split}
&\mathcal{L}_{epl} = \sum_{n_{B}} \sum_{h} \sum_{w} - \tau_{\hat{\mathbf{y}}_{n,h,w}^{t, pr}} M_{n_{B},h,w}^{t}\\
&\qquad\qquad\quad\; sim(F(\mathbf{x}_{n,h,w}^{t} ), \mu_{\hat{\mathbf{y}}_{n,h,w}^{t, pr}}^{s})
\end{split}
\end{align}
where $\mathcal{L}_{ls}$ is the lovasz softmax from \cite{berman2018lovasz, cortinhal2021salsanext}, and is used to maximize IoU. $\mathcal{L}_{wce}$ is the cross entropy loss on the source data, whose pixels are weighted by the reciprocals of the portion of their class labels. $\mathcal{L}_{epl}$ is a loss function that aligns the domains by increasing the similarity between source class prototypes and selectively pseudo labeled target features. $\lambda$ is a hyperparameter for balancing the loss functions.

\section{Experiments}
We evaluate the performance of our method on the recent adaptation scenarios proposed by \cite{yi2021complete}. The original scenario validated six adaptation setups across three datasets, SemanticKITTI~\cite{behley2019semantickitti}, nuScenes-lidarseg~\cite{caesar2020nuscenes} and Waymo Open Dataset~\cite{sun2020waymo}. Among the proposed scenarios, we evaluate our method on the adaptation scenarios between SemanticKITTI and nuScenes-lidarseg, because Waymo does not contain many classes and dense annotations. The performance is evaluated on the commonly used validation split of each dataset. For tables and experiments, we abbreviate SemanticKITTI as \textbf{K} and nuScenes as \textbf{N}.

\begin{table}[t]
\centering
\caption{Experiment results of UDA on the LiDAR semantic segmentation scenarios proposed by \cite{yi2021complete}.}
\begin{tabular}{c|c|ccc}
\toprule
\multicolumn{2}{c|}{Scenario} & K2N  & N2K  & mean \\
\midrule
\midrule
\multirow{6}{*}{3D} & Source Only~\cite{yi2021complete} & 27.9 &23.5 & 25.7\\
                    & FeaDA~\cite{yi2021complete}       & 27.2 &21.4 & 24.3\\
                    & OutDA~\cite{yi2021complete}       & 26.5 &22.7 & 24.6\\
                    & SWD~\cite{yi2021complete}         & 27.2 &24.5 & 25.9\\
                    & 3DGCA~\cite{yi2021complete}       & 27.4 &13.4 & 20.4\\
                    & CnL~\cite{yi2021complete}         & 31.6 &33.7 & 32.7\\
\midrule 
2D & SQSGV2~\cite{yi2021complete, Wu2019SqueezeSegV2IM}      & 10.1 & 23.9 & 17.0\\
\midrule
\midrule
2D & Ours & \textbf{35.8} & \textbf{34.1} & \textbf{35.0}\\
\bottomrule
\end{tabular}

\label{tab:cnluda}
\vspace{-2mm}
\end{table}

\begin{table*}[t]
\footnotesize
\renewcommand{\arraystretch}{0.9}
\centering
\caption{Class-wise experiment results of UDA on the LiDAR semantic segmentation scenarios proposed by \cite{rochan2021unsupervised}. Numbers denote IoU.}
\begin{tabular}{c|c|cccccccccc|c}
\toprule
             Method & Scenario & car  & bcycl & mcycl & othvhc & pedest & truck & drvbl & sdwlk & trrn & vgtn  & mIoU \\
\midrule
\midrule
CORAL+GA \cite{rochan2021unsupervised}    & K2N      & 51.0   & 0.9   & 6.0     & 4.0      & 25.9   & 29.9  & 82.6  & 27.1  & 27.0   & 55.3 & 31.0     \\
GA \cite{rochan2021unsupervised}           & K2N      & 54.4 & 3.0     & 1.9   & 7.6    & 27.7   & 15.8  & 82.2  & 29.6  & 34.0   & 57.9 & 31.4     \\
\midrule
\midrule
Ours         & K2N      &43.7 &1.5 &22.1 &40.6 &17.7 &32.9 &61.6 &29.3 &29.4 &79.4  &\textbf{35.8}      \\
% \bottomrule
% \toprule
\midrule
CORAL+GA \cite{rochan2021unsupervised}    & N2K      & 47.3 & 10.4  & 6.9   & 5.1    & 10.8   & 0.7   & 24.8  & 13.8  & 31.7 & 58.8 & 21.0     \\
GA \cite{rochan2021unsupervised}           & N2K      & 49.6 & 4.6   & 6.3   & 2.0    & 12.5   & 1.8   & 25.2  & 25.2  & 42.3 & 43.4 & 21.3     \\
\midrule
\midrule
Ours         & N2K      &36.4 &9.8 &9.5 &2.3 &4.7 &12.1 &81.7 &55.3 &53.1 &76.0       & \textbf{34.1}    \\
\bottomrule
\end{tabular}
\label{tab:gauda}
\vspace{-2mm}
\end{table*}

\begin{table*}[t]
\centering
\footnotesize
\caption{Ablation results of our method.}
\renewcommand{\arraystretch}{0.9} 
\begin{tabular}{c|ccccccc|c}

\toprule
\multirow{2}{*}{Scenario} & \multirow{2}{*}{\shortstack{Source\\ First}} & \multirow{2}{*}{\shortstack{Recon\\pre-training}} & \multirow{2}{*}{\shortstack{Enhanced\\Prototypes}} & \multirow{2}{*}{\shortstack{Averaged\\Prototypes}} & \multirow{2}{*}{\shortstack{Deactivate\\Mask}} & \multirow{2}{*}{\shortstack{Confidence\\Weighting}} & \multirow{2}{*}{\shortstack{Background\\Down-weighting}} & \multirow{2}{*}{mIoU} \\
                          &                         &                            &                            &                             &                             &                             &                       \\
\midrule
\midrule
K2N      & $\checkmark$ & $\cdot$      & $\checkmark$ & $\checkmark$ & $\checkmark$ & $\checkmark$ & $\checkmark$ & \textbf{30.0}\\
K2N      & $\checkmark$ & $\checkmark$ & $\cdot$      & $\checkmark$ & $\checkmark$ & $\checkmark$ & $\checkmark$ & \textbf{29.7}\\
K2N      & $\checkmark$ & $\checkmark$ & $\checkmark$ & $\cdot$      & $\checkmark$ & $\checkmark$ & $\checkmark$ & 34.4\\
K2N      & $\checkmark$ & $\checkmark$ & $\checkmark$ & $\checkmark$ & $\cdot$      & $\checkmark$ & $\checkmark$ & \textbf{27.9}\\
K2N      & $\checkmark$ & $\checkmark$ & $\checkmark$ & $\checkmark$ & $\checkmark$ & $\cdot$      & $\checkmark$ & \textbf{30.3}\\
K2N      & $\checkmark$ & $\checkmark$ & $\checkmark$ & $\checkmark$ & $\checkmark$ & $\checkmark$ & $\cdot$      & 34.9\\
K2N      & $\checkmark$ & $\checkmark$ & $\checkmark$ & $\checkmark$ & $\checkmark$ & $\checkmark$ & $\checkmark$ & 35.8\\
\midrule
N2K      & $\cdot$ & $\checkmark$ & $\checkmark$ & $\checkmark$ & $\checkmark$ & $\checkmark$ & $\checkmark$ & \textbf{29.3}\\
N2K      & $\checkmark$ & $\checkmark$ & $\checkmark$ & $\checkmark$ & $\checkmark$ & $\checkmark$ & $\checkmark$ & 34.1\\
\bottomrule
\end{tabular}

\label{tab:ablation}
\vspace{-2mm}
\end{table*}

\subsection{Scenario details}
Since the original paper~\cite{yi2021complete} of the proposed scenarios does not provide accurate mappings between scenarios, we have read through the documents of each dataset and implemented the scenarios on our own. The scenarios are evaluated on ten classes, \{car, bicycle, motorcycle, other vehicle, pedestrian, truck, drivable surface, sidewalk, terrain, vegetation\}. We only elaborate the mappings between classes that are confusing. On the SemanticKITTI dataset, we have mapped \{bus, other vehicle\} to \{other vehicle\}, \{road, lane marking\} to \{drivable surface\}, and \{vegetation, trunk\} to \{vegetation\}. The moving classes are mapped in accordance with their non-moving class counterparts. On nuScenes,  \{adult, child, construction worker, police officer\} were mapped to \{pedestrian\}, and \{bus, construction vehicle, trailer\} were mapped to \{other vehicle\}.

\subsection{Implementation details}
We implemented our method using the PyTorch framework, and four NVIDIA RTX 2080 Ti GPUs were used to train our model. Our model is trained and inferred using the FP16 (half-precision) setup of PyTorch. Every iteration, we have sampled 4 samples from the source domain and 4 samples from the target domain to train our network. For optimization, we have used the Stochastic Gradient Descent (SGD) with learning rate 0.01, momentum 0.9 and weight decay 0.0001. For the pre-training stage, we used learning rate warmup, and linearly increased the learning rate from initial value 0.0001 to 0.01 during the first epoch. After warmup, we used the exponential scheduler, which decays the learning rate at every epoch with gamma 0.99. For the joint training stage, we did not use warmup and used an inverse scheduler that is widely used in domain adaptation setups \cite{ganin2015unsupervised}. The learning rate is scheduled with $\eta_{p}=\eta_{0}/(1+ap)^{b}$ with $\eta_{0}$ the initial learning rate, $a$ and $b$ hyperparameters which are 10 and 0.75 each, and $p$ a value that linearly increases from $0$ to $1$ iteration-wise. We have pre-trained our networks for 50 epochs, and performed joint training for 30 epochs. The $\alpha$ in (9) is set to 0.99, the $p_{inc}$ used for calculating the $\tau_{c}$ in (10) is set to 0.01, and the $p_{dw}$ used in down-weighting class $0$ is set to 0.1. The $\lambda$ in (11) is set to 1.0, and we did not finetune it further.

Following former projection-based methods \cite{milioto2019rangenet}, we use KNN post-processing, which propagates the 2D predictions to 3D neighbors. Details can be found in \cite{milioto2019rangenet}. The required parameters are $\{S, k, \text{cutoff}, \delta\}$ and we have used $\{5, 5, 1.0, 1.0\}$, each, following the original setup.

\subsection{Comparison against contemporary methods}
On Table 2 and Table 3 we compare our method against contemporary methods. Direct comparison against the two methods is inaccurate, as the detailed settings of \cite{yi2021complete} are unknown, and the settings of \cite{rochan2021unsupervised} are slightly different from ours. Still it can be seen that our method outperforms contemporary methods by a meaningful margin. Regarding \cite{yi2021complete}, the improvement over their method is meaningful, because 3D methods usually outperform 2D methods due to the richer input representation. Against \cite{rochan2021unsupervised}, the gap on the N2K scenario is noticeable, and it is also meaningful that the performance of our method is not imbalanced across scenarios.

\subsection{Ablation studies on method components}
On Table 4 we show the performance of our model with certain components ablated. The performance of the five components which contribute to a large mIoU decrease are denoted in bold letters. The N2K experiment without Source First Principle corresponds to a setup where the nuScenes data have been upsampled via nearest neighbor interpolation. The large mIoU reduction supports our claim that keeping the source representation as intact as possible is crucial. Mask deactivation, despite its simplicity, shows a large mIoU decrease if ablated. This shows that it is important to mitigate the learning difficulties induced from the inclusion of class $0$. When compared against background down-weighting, it can be seen that mask deactivation is a stronger component for dealing with class $0$. When averaged prototypes are ablated, we obtain the prototypes from the classes in the current source scene only. The small performance decrease of this setup means that the performance of our method is mainly maintained by classes that are prevalent in every scene.

\subsection{Runtime evaluation}
We evaluate the inference time of our method on a single RTX 2080 Ti, with CUDA synchronization for accurate time measurement and FP16 activated. On the K2N scenario, it operates at the speed of 56 FPS (CNN: 16.50 ms, KNN: 1.36 ms per image) and on the N2K scenario it operates at the speed of  32 FPS (CNN: 28.16 ms, KNN: 2.63 ms per image) which is a speed that supports real-time applications.

\subsection{Limitations and future directions}
In the scenarios proposed by \cite{yi2021complete} where the Waymo dataset is included, it is difficult to produce meaningful results, because the number of class $0$ pixels are excessively large. Modifying the model design to attend more on the semantic classes, and discriminating class $0$ pixels more aggressively could be a future research direction to pursue. Also, as stated in the ablation experiments, our method is focused on the prevalent classes, and could be further improved by attending on the minority classes.
\subsection{Acknowledgements}
This research was supported by the Engineering Research Center Program through the National Research Foundation of Korea (NRF) funded by the Korean Government MSIT (NRF-2018R1A5A1059921) and by the Institute of Information \& communications Technology Planning \& Evaluation (IITP) grant funded by the Korea government(MSIT) (No.2021-0-02068, Artificial Intelligence Innovation Hub).

\bibliography{cite}

\end{document}